\documentclass[10pt,twocolumn,letterpaper]{article}

\usepackage{cvpr}
\usepackage{times}
\usepackage{epsfig}
\usepackage{graphicx}
\usepackage{amsmath}
\usepackage{amssymb}
\usepackage{caption}
\usepackage{color}
\usepackage{array}
\usepackage{lipsum}

\newcommand{\lqAdd}[1]{{\color{red}#1}}
\newcommand{\lqDelete}[1]{{\color{cyan}#1}}

\newcommand{\CQ}[1]{{\color{black}#1}}

\cvprfinalcopy 

\def\cvprPaperID{4318} 

\begin{document}

\title{SketchyCOCO: Image Generation from Freehand Scene Sketches}



\author{Chengying Gao$^1$ \quad Qi Liu$^1$ \quad Qi Xu$^1$ \quad Limin Wang$^2$ \quad Jianzhuang Liu$^{3}$ \quad Changqing Zou $^{4}$\thanks{Corresponding author.} \\
$^1$School of Data and Computer Science, Sun Yat-sen University, China \\ 
$^2$State Key Laboratory for Novel Software Technology, Nanjing University, China \\
$^3$Noah's Ark Lab, Huawei Technologies \quad $^4$HMI Lab, Huawei Technologies \\
{\tt\small mcsgcy@mail.sysu.edu.cn \{liuq99, xuqi5\}@mail2.sysu.edu.cn}\\
{\tt\small 07wanglimin@gmail.com  liu.jianzhuang@huawei.com aaronzou1125@gmail.com}
}

\maketitle
\thispagestyle{empty}

\begin{abstract}
We introduce the first method for automatic image generation from scene-level freehand sketches. Our model allows for controllable image generation by specifying the synthesis goal via freehand sketches. The key contribution is an attribute vector bridged Generative Adversarial Network called EdgeGAN, which supports high visual-quality object-level image content generation without using freehand sketches as training data. We have built a large-scale composite dataset called SketchyCOCO to \CQ{support and evaluate} the solution. We validate our approach on the {tasks} of both object-level and scene-level image generation on SketchyCOCO. Through quantitative, qualitative results, human evaluation and ablation studies, we demonstrate the method's capacity to generate realistic complex scene-level images from various freehand sketches. 
\end{abstract}

\section{Introduction}
In recent years Generative Adversarial Networks (GANs) \cite{NIPS2014} have shown significant success in modeling high dimensional distributions of visual data. In particular, high-fidelity images could be achieved by unconditional generative models trained on object-level data (e.g., animal pictures in \cite{brock2018large}), class-specific datasets (e.g., indoor scenes~\cite{Silberman2012}), or even a single image with repeated textures~\cite{shaham2019singan}. For practical applications, automatic image synthesis which can generate images and videos in response to specific requirements could be more useful. This explains why there are increasingly studies on the adversarial networks conditioned on another input signal like texts~\cite{zhang2017stackgan,HongYCL18}, semantic maps~\cite{ashual2019specifying, pix2pix2017,chen2017photographic,turkoglu2019layer,park2019SPADE}, layouts~\cite{ashual2019specifying, HongYCL18,ZhaoMYS19}, and scene graphs \cite{ashual2019specifying,johnson2018image}.  Compared to these sources, a freehand sketch has its unique strength in expressing the user's idea in an intuitive and flexible way. Specifically, to describe an object or scene, sketches can better convey the user's intention than other sources since they lessen the uncertainty by naturally providing more details such as object location, pose and shape.  

\begin{figure*}   
\footnotesize
\begin{minipage}{0.12\textwidth}  
  \centerline{\includegraphics[width=1\textwidth]{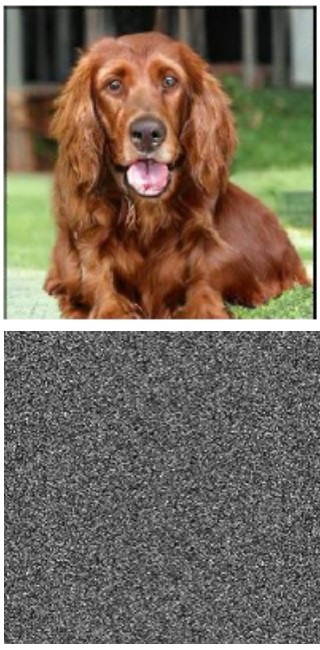}}
  \centerline{BigGAN~\cite{brock2018large}}
\end{minipage}
\hfill
\begin{minipage}{0.12\textwidth}  
  \centerline{\includegraphics[width=1\textwidth]{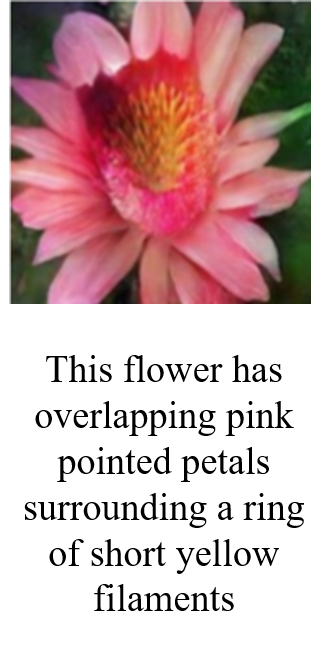}}
  \centerline{StackGAN~\cite{zhang2017stackgan}}
\end{minipage}
\hfill
\begin{minipage}{0.12\textwidth}
  \centerline{\includegraphics[width=1\textwidth]{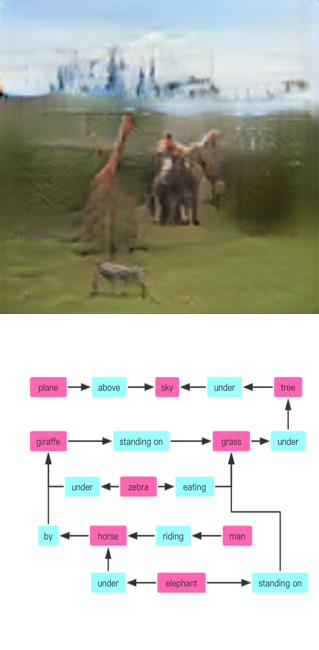}}
  \centerline{Sg2im\cite{johnson2018image}}
\end{minipage}
\hfill
\begin{minipage}{0.12\textwidth}
  \centerline{\includegraphics[width=1\textwidth]{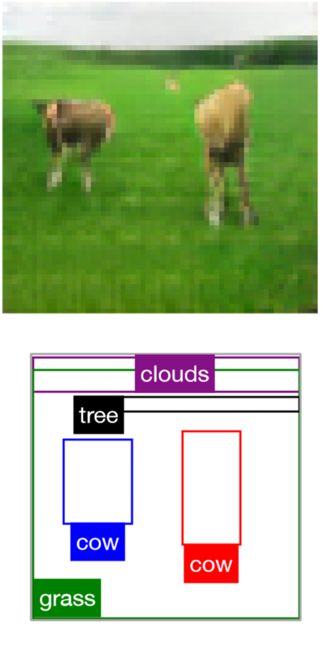}}
  \centerline{Layout2im~\cite{ZhaoMYS19}}
\end{minipage}
\begin{minipage}{0.12\textwidth}  
  \centerline{\includegraphics[width=1\textwidth]{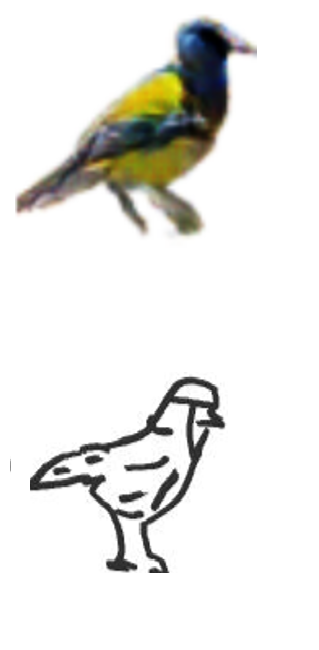}}
  \centerline{ContexturalGAN~\cite{lu2018image}}
\end{minipage}
\hfill
\begin{minipage}{0.12\textwidth}
  \centerline{\includegraphics[width=1\textwidth]{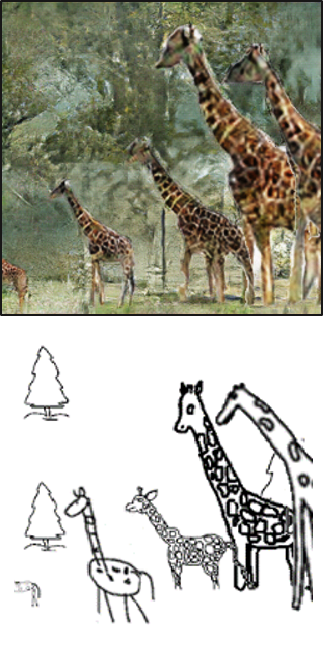}}
  \centerline{Ours}
\end{minipage}
\hfill
\begin{minipage}{0.12\textwidth}
  \centerline{\includegraphics[width=1\textwidth]{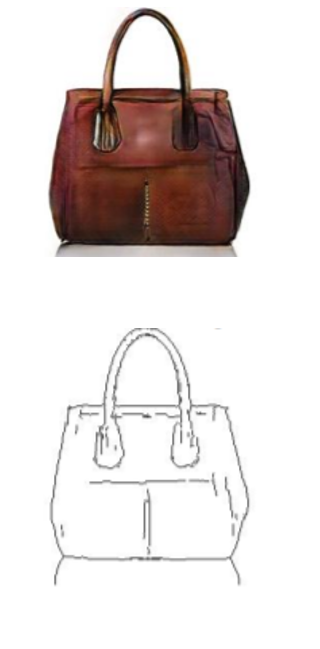}}
  \centerline{Pix2pix~\cite{pix2pix2017}}
\end{minipage}
\hfill
\begin{minipage}{0.12\textwidth}
  \centerline{\includegraphics[width=1\textwidth]{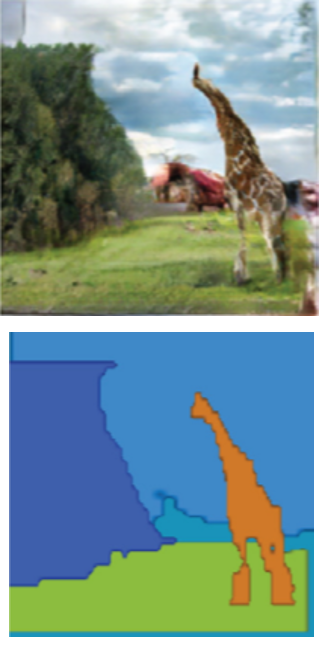}}
  \centerline{Ashual et al.~\cite{ashual2019specifying}}
\end{minipage}
\caption{The proposed approach allows users to controllably generate realistic scene-level images with many objects from freehand sketches, which is in stark contrast to unconditional GAN and conditional GAN in that we use scene sketch as context (a weak constraint) instead of generating from noise~\cite{brock2018large} or with harder condition like semantic maps~\cite{ashual2019specifying,park2019semantic} or edge maps~\cite{pix2pix2017}.
\CQ{The constraints of input become stronger from left to right.} 
}
\label{fig:teaser}
\normalsize
\end{figure*}

In this paper, we extend the use of Generative Adversarial Networks into a new problem: controllably generating realistic images with many objects and relationships from a freehand scene-level sketch as shown in Figure~\ref{fig:teaser}. This problem is extremely challenging because of several factors. 
Freehand sketches are characterized by various levels of abstractness, for which there are a thousand different appearances from a thousand users, which even express the same common object, {depending on the users' depictive abilities}, thereby making it difficult for existing techniques to model the mapping from a freehand scene sketch to realistic natural images that precisely meet the users' intention. 
More importantly, freehand scene sketches are often incomplete and contain a foreground and background. For example, users often prefer to sketch the foreground object, which are most concerned, with specific detailed appearances and they would like the result to exactly {satisfy} this requirement while 
they leave blank space and just draw the background objects roughly without paying attention to their details,
thereby requiring the algorithm to be capable of coping with the different requirements of users.

To make this challenging problem resolvable, we decompose {it} into two sequential stages, foreground and background generation, based on the characteristics of scene-level sketching. The first stage focuses on foreground generation where the generated image content 
is supposed to exactly meet the user's specific requirement. The second stage is responsible for background generation where the generated image content may be loosely aligned with the sketches. Since the appearance of each object in the foreground has been specified by the user, it is possible to generate realistic and reasonable image content from the individual foreground objects separately. Moreover, the generated foreground can provide more constraints on the background generation, which makes background generation easier, i.e., progressive scene generation reduces the complexity of the problem.

To address the data variance problem caused by the abstractness of sketches, we propose a new neural network architecture called EdgeGAN. It learns a joint embedding to transform images and the corresponding various-style edge maps into a shared latent space in which vectors can represent high-level attribute information (i.e., object pose and appearance information) from cross-domain data. With the bridge of the attribute vectors in the shared latent space, we are able to transform the problem of image generation from freehand sketches to the one from edge maps without the need to collect foreground freehand sketches as training data, and we can  address the challenge of modeling one-to-many correspondences between an image and infinite freehand sketches.

To evaluate our approach, we build a large-scale composite dataset called SketchyCOCO based on MS COCO Stuff~\cite{caesar2016coco}. {The current version of this dataset includes 14K+ pairwise examples of scene-level images and sketches, 20K+ triplet examples of foreground sketches, images, and edge maps which cover 14 classes, 27K+ pairwise examples of background sketches and image examples which cover 3 classes, and the segmentation ground truth of 14K+ scene sketches. We compare the proposed EdgeGAN to existing sketch-to-image approaches. Both qualitative and quantitative results show that the proposed EdgeGAN achieves significantly superior performance.

{We summarize our contributions as follows:
\begin{itemize}
\item We propose the first deep neural network based framework for image generation from  scene-level freehand sketches.
\item We contribute a novel generative model called EdgeGAN for object-level image generation from freehand sketches. This model can be trained in an end-to-end manner and does not require sketch-image pairwise ground truth for training. 
\item We construct a large scale composite dataset called SketchyCOCO based on MS COCO Stuff~\cite{caesar2016coco}. This dataset will greatly facilitate related research.
\end{itemize}
}

\section{Related Work}
\textbf{{Sketch-Based Image Synthesis.}}
Early sketch-based image synthesis approaches are based on image retrieval. Sketch2Photo~\cite{chen2009sketch2photo} and PhotoSketcher~\cite{eitz2011photosketcher} synthesize realistic images by compositing objects and backgrounds retrieved from a given sketch.  PoseShop~\cite{chen2013poseshop} composites images of people by letting users input an additional 2D skeleton into the query so that the retrieval will be more precise. Recently, SketchyGAN~\cite{chen2018sketchygan} and ContextualGAN~\cite{lu2018image} have demonstrated the value of variant GANs for image generation from freehand sketches. Different from SketchyGAN~\cite{chen2018sketchygan} and ContextualGAN~\cite{lu2018image}, which mainly solve the problem of image generation from object-level sketches depicting single objects, our approach focuses on generating images from scene-level sketches.  

\textbf{{Conditional Image Generation.}}  Several recent studies have demonstrated the potential of variant GANs for scene-level complex image generation from text~\cite{zhang2017stackgan,HongYCL18}, scene graph~\cite{johnson2018image}, semantic layout map~\cite{HongYCL18,ZhaoMYS19}.
 Most of these methods use a multi-stage coarse-to-fine strategy to infer the image appearances of all semantic layouts in the input or intermediate results at the same time. We instead take another way and use a divide-and-conquer strategy to sequentially generate the foreground and background appearances of the image 
because of the unique characteristics of freehand scene sketches where foreground and background are obvious different.

On object-level image generation, our EdgeGAN is in stark contrast to unconditional GANs and conditional GANs in that we use a sketch as context (a weak constraint) instead of generating from noise like DCGAN~\cite{radford2015unsupervised}, Wasserstein GANs~\cite{arjovsky2017wasserstein}, WGAN-GP~\cite{gulrajani2017improved} and their variants, or with hard condition such as an edge map~\cite{cheng2015deep, deshpande2015learning, LarssonMS16, pix2pix2017}, semantic map~\cite{ashual2019specifying, pix2pix2017,chen2017photographic,turkoglu2019layer,park2019SPADE}, while providing more precise
control than those using text~\cite{zhang2017stackgan,HongYCL18}, layout~\cite{ashual2019specifying, HongYCL18,ZhaoMYS19} and scene graph~\cite{ashual2019specifying,johnson2018image} as context.

\begin{figure}[]
	\centering  
	\includegraphics[width=1.0\linewidth]{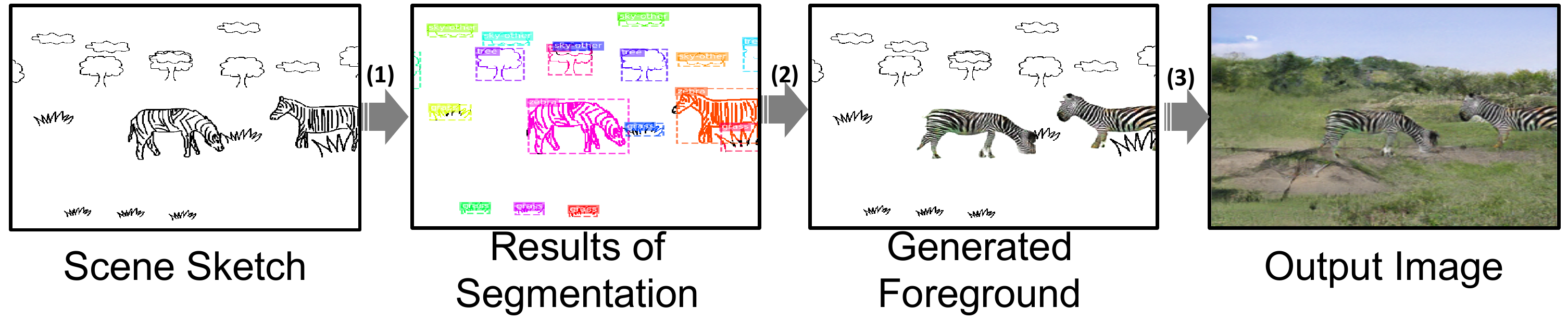}
	\caption{Workflow of the proposed framework.} 
	\label{fig:workflow}
\vspace{-0.5em}
\end{figure}

\section{Method}
Our approach mainly includes \CQ{two} sequential modules: foreground generation and background generation. 
As illustrated in Fig.~\ref{fig:workflow}, given a scene sketch, the object instances are first located and recognized by leveraging the sketch segmentation method in \cite{ZouMGDF19}. After that image content is generated for each foreground object instance (i.e., sketch instances belonging to the foreground categories) individually in a random order by the foreground generation module. By taking background sketches and the generated foreground image as input, the final image is achieved by generating the background image in a single pass. \CQ{The two modules are trained separately.} We next describe the details of each module. 

\begin{figure*}[]
	\centering  
	\includegraphics[width=1.0\linewidth]{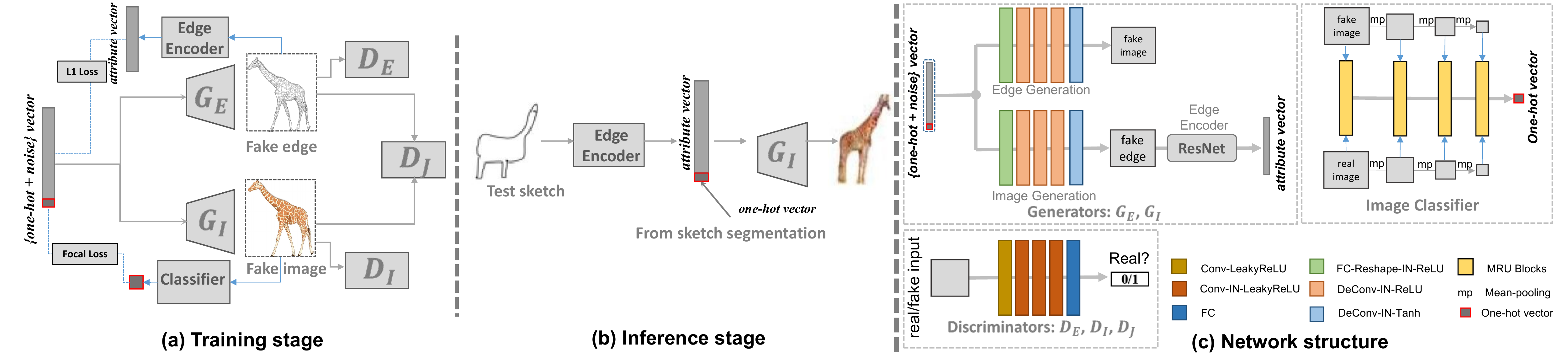}
	\caption{Structure of the proposed EdgeGAN. It contains four sub-networks: two generators $G_I$ and $G_E$, three discriminators $D_I$, $D_E$, and $D_J$, an edge encoder $E$ and an image classifier $C$.
	{EdgeGAN} learns a joint embedding for an image and various-style edge maps depicting this image into a shared latent space where vectors can encode high-level attribute information from cross-modality data.}%
	\label{fig:edgeGAN}
\end{figure*}

\subsection{Foreground Generation}
\label{subsec:instGen}
\noindent\textbf{Overall Architecture of EdgeGAN.}
Directly modeling the mapping between a single image and its corresponding sketches, such as SketchyGAN~\cite{chen2018sketchygan}, is difficult because of the enormous size of the mapping space. We therefore instead address the challenge in another feasible way instead: we learn a common representation for an object expressed by cross-domain data. 
To this end, we design an adversarial architecture, which is shown in Fig.~\ref{fig:edgeGAN}(a), for EdgeGAN. Rather than directly inferring images from sketches, EdgeGAN transfers the problem of sketch-to-image generation to the problem of generating the image from an attribute vector that is encoding the expression intent of the freehand sketch. 
At the training stage, EdgeGAN learns a common attribute vector for an object image and its edge maps by feeding adversarial networks with images and their various-drawing-style edge maps. 
At the inference stage (Fig.~\ref{fig:edgeGAN} (b)), EdgeGAN captures the user's expression intent with an attribute vector and then generates the desired image from it.

\noindent\textbf{Structure of EdgeGAN.}
As shown in Fig.~\ref{fig:edgeGAN}(a), the proposed EdgeGAN has two channels: one including generator $G_E$ and discriminator $D_E$ for edge map generation, the other including generator $G_I$ and discriminator $D_I$ for image generation. Both $G_I$ and $G_E$ take the same
noise vector together with an one-hot vector indicting a specific category as input.
Discriminators $D_I$ and $D_E$ attempt to distinguish the generated images or edge maps from real distribution. 
 Another discriminator $D_J$ is used to encourage the generated fake image and the edge map depicting the same object by telling if the generated fake image matches the fake edge map, which takes the outputs of both $G_I$ and $G_E$ as input (the image and edge map are concatenated along the 
 {width} dimension). The Edge Encoder is used to encourage the encoded attribute information of edge maps to be close to the noise vector fed to $G_I$ and $G_E$ through a $L1$ loss. The classifier is used to infer the category label of the output of $G_I$, which is used to encourage the generated fake image to be recognized as the desired category via a focal loss~\cite{lin2018focal}. The detailed structures of each module of EdgeGAN are illustrated in Fig.~\ref{fig:edgeGAN}(c). 

We implement the Edge Encoder with the same encoder module in bicycleGAN~\cite{zhu2017toward} since they play a similar role functionally, i.e., our encoder encodes the ``content" (e.g., the pose and {shape} information), while the encoder in bicycleGAN encodes properties into latent vectors. For Classifier, we use an architecture similar to the discriminator of SketchyGAN while ignoring the adversarial loss and only using the focal loss~\cite{lin2018focal} as the classification loss. The architecture of all generators and discriminators are based on WGAP-GP~\cite{gulrajani2017improved}. Objective function and more training details can be found in the supplementary materials.

\subsection{Background Generation}
\label{subsec:backGen}
Once all of the foreground instances have been synthesized, we train pix2pix~\cite{pix2pix2017} to generate the background. The major challenge of the background generation task is that 
the background of most scene sketches contains both the background instance and the blank area within the area(as shown in Fig.~\ref{fig:workflow}), which means some area belonging to the background is uncertain because of the lack of sketch constraint. 
By leveraging pix2pix and using the generated foreground instances as constraints, we can allow the network to generate a reasonable background matching the synthesized foreground instances. Taking Fig.~\ref{fig:workflow} as an example, the region below the zebras of the input image contains no background sketches for constraints, 
and the output image shows that such a region can be reasonably filled in with grass and ground.

\section{SketchyCOCO Dataset}

\begin{figure*}[h]
	\centering  
	\includegraphics[width=1.0\linewidth]{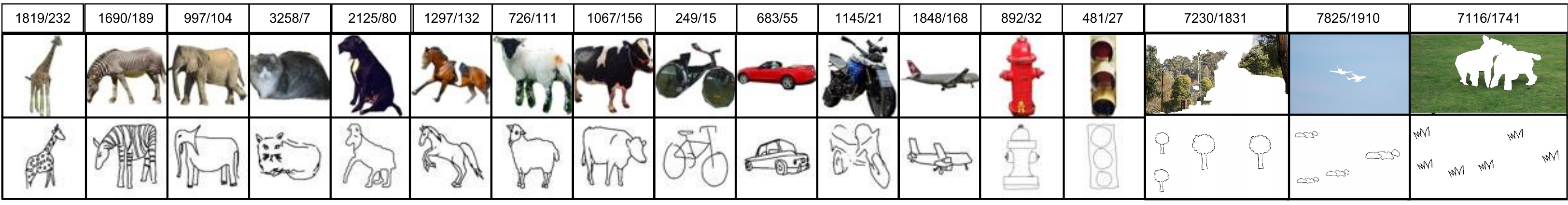}
	\caption{Representative sketch-image pairwise examples from 14 foreground and 3 background categories in SketchyCOCO. The data size of each individual category,  splitting to training/test, is shown on the top.}
	\label{fig:objdb}
\end{figure*}

\begin{figure*}[]
	\centering  
	\includegraphics[width=0.96\linewidth]{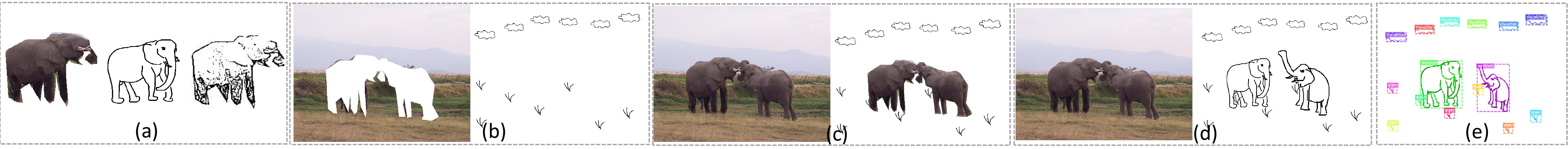}
	\caption{Illustration of five-tuple ground truth data of SketchyCOCO, i.e., (a) \{foreground image, foreground sketch, foreground edge maps\} (training: 18,869, test: 1,329), (b) \{background image, background sketch\} (training: 11,265, test: 2,816), (c) \{scene image, foreground image \& background sketch\} (training: 11,265, test: 2,816), (d) \{scene image, scene sketch\} (training: 11,265, test: 2,816), and (e) sketch segmentation (training: 11,265, test: 2,816). 
	}
	\label{fig:9tuple}
	\vspace{-0.5em}
\end{figure*}

We initialize the construction
by collecting instance freehand sketches covering 3 background classes and 14 foreground classes from the Sketchy dataset~\cite{sangkloy2016sketchy}, Tuberlin dataset~\cite{eitz2012humans}, and QuickDraw dataset~\cite{ha2017neural} (\CQ{around 700 sketches for each foreground class}). For each class, we split these sketches into two parts: $80\%$ for the training set, and the remaining $20\%$ for the test set.
We collect 14081 natural images from COCO Stuff~\cite{caesar2016coco} containing at least one of 17 categories and split them into two sets, $80\%$ for training and the remaining $20\%$ for test.
Using the segmentation masks of these natural images, we place background instance sketches (clouds, grass, and tree sketches) at random positions within the corresponding background regions of these images.
This step produces $27,683 (22,171+5,512)$ pairs of background sketch-image examples (shown in  Fig.~\ref{fig:objdb}). 

After that, for each foreground object in the natural image, we retrieve the most similar sketch with the same class label as the corresponding foreground object in the image. 
This step employs the sketch-image embedding method proposed in the Sketchy database~\cite{sangkloy2016sketchy}. In addition, in order to obtain more data for training object generation model, we collect foreground objects from the full COCO Stuff dataset. With this step and the artificial selection, we obtain $20,198(18,869+1,329)$ triplets examples of foreground sketches, images 
and edge maps. Since all the background objects and foreground objects of natural images from COCO Stuff have category and layout information, we therefore obtain the layout (e.g., bounding boxes of objects) and segmentation information for the synthesized scene sketches as well. After the construction of both background and foreground sketches, we naturally obtain five-tuple ground truth data~(Fig.~\ref{fig:9tuple}). Note that in the above steps, scene sketches in training and test set can only be made up by instance sketches from the training and test sets, respectively.

\section{Experiments}
\subsection{Object-level Image Generation}
\label{subsec:abla}
\noindent\textbf{Baselines.} 
We compare EdgeGAN with the general image-to-image model {pix2pix}~\cite{pix2pix2017} and two existing sketch-to-image models,  {ContextualGAN}~\cite{lu2018image} and {SketchyGAN}\cite{chen2018sketchygan}, on the collected 20,198 triplets \{foreground sketch, foreground image, foreground edge maps\} examples.
Unlike SketchyGAN and pix2pix which may use both edge maps and freehand sketches for training data, 
EdgeGAN and ContextualGAN take as input only edge maps and do not use any freehand sketches for training. For fair and thorough evaluation, we set up several different training modes for SketchyGAN, pix2pix, and ContextualGAN. We next introduce these modes for each model.


\begin{itemize}
\item \noindent\textbf{EdgeGAN:} we train a \CQ{single} model using foreground images and only {the extracted} edge maps for all 14 foreground object categories. 

\item \noindent\textbf{ContextualGAN ~\cite{lu2018image}:} we use foreground images and their edge maps to separately train a model for each foreground object category, since the original method cannot use a single model to learn the sketch-to-image correspondence for multiple categories. 

\item \noindent\textbf{SketchyGAN~\cite{chen2018sketchygan}:} 
we train the original SketchyGAN in two modes. The first mode denoted as SketchyGAN-{E} uses foreground images and only their edge maps for training. Since SketchyGAN may use both edge maps and freehand sketches for training data in their experiments, we also train SketchyGAN in another mode: using foreground images and \{their edge maps + sketches\} for training. In this training mode called SketchyGAN-{E\&S}, we follow the same training strategy as SketchyGAN did to feed edge maps to the model first and then fine-tune it with sketches. 

\item \noindent\textbf{pix2pix ~\cite{pix2pix2017}:}
we train the original pix2pix architecture in four modes. The first two modes are denoted as  pix2pix-{E}-{SEP} and pix2pix-{S}-{SEP}, in which we separately train 14 models by using only edge maps or sketches from the 14 foreground categories, respectively. The other two modes are denoted as pix2pix-{E}-{MIX} and pix2pix-{S}-{MIX}, in which we train a single model respectively using only edge maps or sketches from all 14 categories.

\end{itemize}

\noindent\textbf{Qualitative results.} 
We show the representative results of the \CQ{four} comparison methods in Fig~\ref{fig:instance_comparison}. In general, EdgeGAN provides much more realistic results than ContextualGAN. In terms of the faithfulness (i.e., whether the input sketches can depict the generated images), EdgeGAN is also superior than ContextualGAN. This can be explained by the fact that EdgeGAN uses the learned attribute vector, which captures reliable high-level attribute information from the cross-domain data for the supervision of image generation. In contrast, ContextualGAN uses a low-level sketch-edge similarity metric for the supervision of image generation, which is sensitive to the abstractness level of the input sketch. 

Compared to EdgeGAN which produces realistic images, pix2pix and SketchyGAN which just colorize the input sketches and do not change the original shapes of the input sketches when the two models are trained with only edge maps (e.g., see Fig.~\ref{fig:instance_comparison} (b1), (c1), and (c2)). This may be because the outputs of both SketchyGAN and pix2pix are strongly constrained by the input (i.e., one-to-one correspondence provided by the training data). When the input is a freehand sketch from another domain, these two models are weak to produce realistic results since they only see edge maps during the training. In contrast, the output of EdgeGAN is relatively weakly constrained by the input sketch since its generator takes as input the attribute vector learnt from cross-domain data rather than the input sketch. Therefore, EdgeGAN can achieve better results than pix2pix and SketchyGAN because it is relatively insensitive to cross-domain input data. 

By augmenting or changing the training data with freehand sketches, \CQ{both SketchyGAN and pix2pix can produce realistic local patches for some categories but fail to preserve
the global shape information, as we can see that the shapes of the results in Fig.~\ref{fig:instance_comparison} (b2), (c3), and (c4) are distorted}.

\begin{figure}[h]
	\centering
	\includegraphics[width=1.0\linewidth]{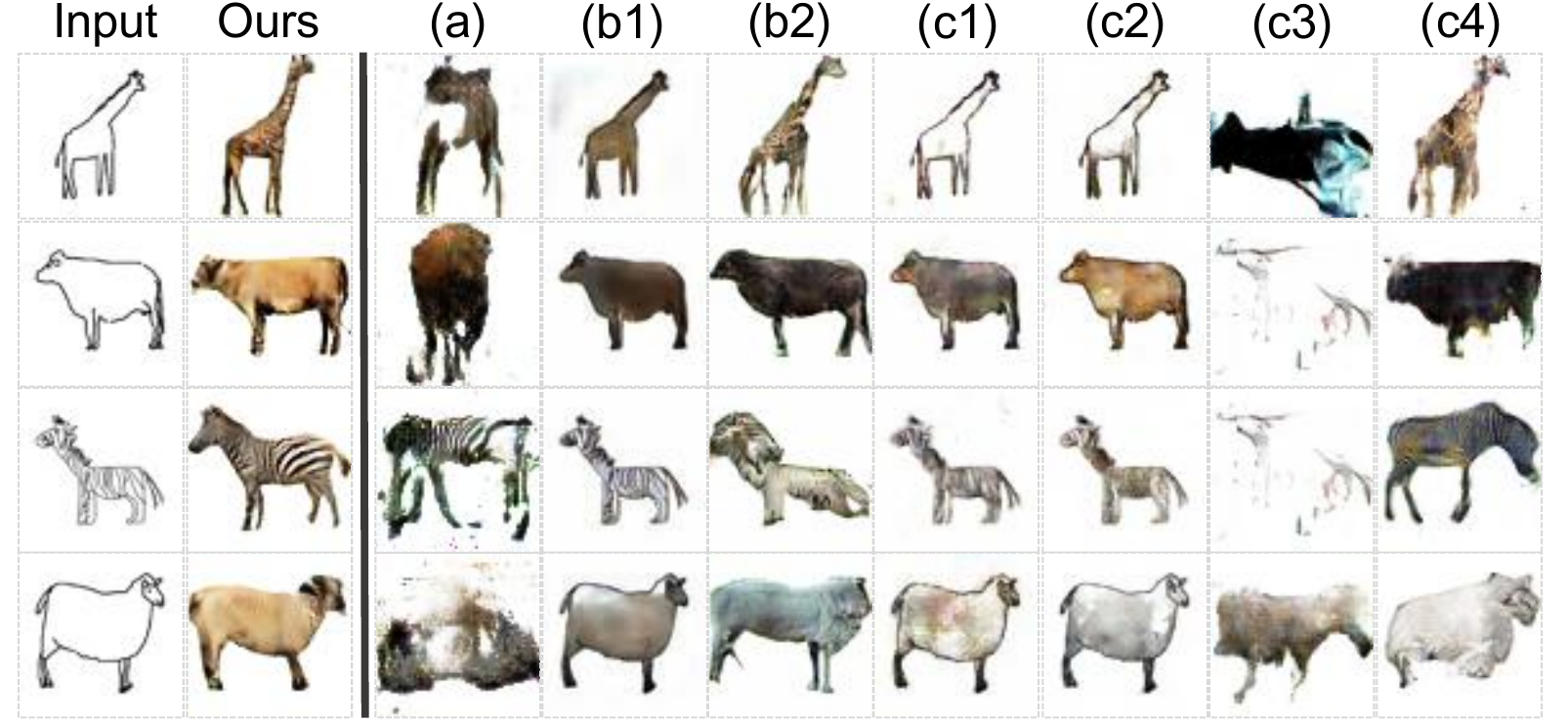}
	\caption{From left to right: input sketches, results from EdgeGAN, ContextualGAN~(a), two training modes of SketchyGAN (i.e., SketchyGAN-E~(b1) and SketchyGAN-E\&S)~(b2), four training modes of pix2pix, i.e,  pix2pix-E-SEP~(c1), pix2pix-E-MIX~(c2),pix2pix-S-MIX(c3), and pix2pix-S-SEP(c4)}	\label{fig:instance_comparison}
\end{figure}

\noindent\textbf{Quantitative results.} 
We carry out both realism and faithfulness evaluations for quantitative comparison. We use FID~\cite{heusel2017gans} and Accuracy~\cite{ashual2019specifying} as the realism metrics. 
Lower FID value and higher accuracy value indicate better image realism.
It is worth mentioning that the Inception Score~\cite{salimans2016improved} metric is not suitable for our task, as several recent researches including~\cite{Borji2019} find the Inception Score is basically only reliable for the models trained on ImageNet. We measure the faithfulness of the generated image by computing \CQ{the extent of the similarity between the edge map of the generated image and the corresponding input sketch}. \CQ{Specifically, we use Shape Similarity (SS), which is the $L2$ Gabor feature~\cite{eitz2012sketch} distance between the input sketch and the edge map generated by the canny edge detector from the generated image, to measure the faithfulness (lower value indicates higher faithfulness)}.



The quantitative results are summarized as Table~\ref{object_score} where we can see that the proposed EdgeGAN achieves the best results
in terms of the realism metrics. However, in terms of the faithfulness metric, our method is better than most of the competitors but is not as good as pix2pix-E-SEP, pix2pix-E-MIX, SketchyGAN-E. This is because the results generated by these methods look more like a colorization of the input sketches whose shapes are almost the same as the input sketch (see Fig.~\ref{fig:instance_comparison} (b1), (c1), (c2)), rather than being realistic. The quantitative results basically confirm our observations in the qualitative study.

\begin{figure*}[]
	\centering  
	\includegraphics[width=0.96\linewidth]{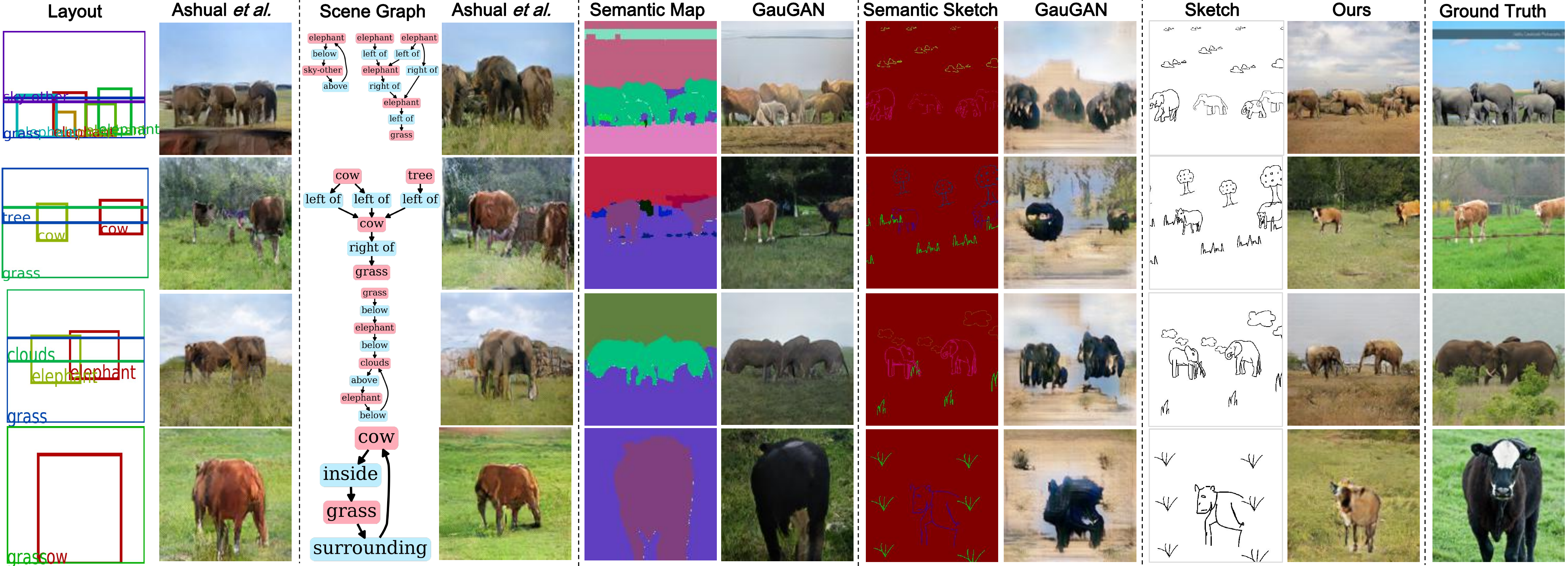}
	\caption{Scene-level comparison. Please see the text in Section 5.2 for the details.} 
	\label{fig:scene_comparison}
\vspace{-1.5em}
\end{figure*}

\newcommand{\tabincell}[2]{\begin{tabular}{@{}#1@{}}#2\end{tabular}}
\begin{table}
\footnotesize
\caption{The results of quantitative experiments and human evaluation.} 
\vspace{-1.5em}
\begin{center}
	\begin{tabular}{c p{0.5cm} p{0.5cm} p{0.6cm} || p{0.5cm} p{0.5cm}}
		\hline
		Model (object) & FID & \tabincell{c}{Acc.} & \tabincell{c}{SS\\(e+04)} & \tabincell{c}{Real-\\ism} & \tabincell{c}{Faith-\\fulness}\\
		\hline
		Ours & \textbf{87.6} & \textbf{0.887} & 2.294 & \textbf{0.637} & 0.576\\
		ContextualGAN & 225.2 & 0.377 & 2.660 & 0.038 & 0.273\\
		SketchyGAN-{E} & 141.5 & 0.277 & \textbf{1.996} & 0.093 &  \textbf{0.945}\\
		SketchyGAN-{E\&S} & 137.9 & 0.127 & 2.315 & 0.023 & 0.691\\
		pix2pix-{E}-{SEP} & 143.1 & 0.613 & 2.136 & 0.071 & 0.918\\
		pix2pix-{E}-{MIX} & 128.8 & 0.499 & 2.103 & 0.058 & 0.889\\
		pix2pix-{S}-{MIX} & 163.3 & 0.223 & 2.569 & 0.047 & 0.353\\
		pix2pix-{S}-{SEP} & 196.0 & 0.458 & 2.527 & 0.033 & 0.310\\
		\hline
		Model (scene) & FID & SSIM & \tabincell{c}{FID\\(local)} & \tabincell{c}{Real-\\ism} & \tabincell{c}{Faith-\\fulness}\\
		\hline
		Ashual et al.~\cite{ashual2019specifying}-layout &123.1 & 0.304 & 183.6 & 0.083 & 1.874\\
		Ashual et al.~\cite{ashual2019specifying}-scene graph & 167.7 & 0.280 & 181.9 & 0.118 & 1.570\\
		GauGAN-semantic map & \textbf{80.3} & \textbf{0.306} & 123.0 & 0.208 & \textbf{2.894}\\
		GauGAN-semantic sketch & 215.1 & 0.285 & 239.5 & 0.000 & 1.210\\
		Ours & 164.8 & 0.288 & \textbf{112.0} & \textbf{0.591} & 2.168\\
		\hline
	\end{tabular}
\label{object_score}
\end{center}
\vspace{-2.5em}
\end{table}

\subsection{Scene-level Image Generation}
\label{subsec:abla} 
\noindent\textbf{Baselines.} 
There is no existing approach which is specifically designed for image generation from scene-level freehand sketches.
SketchyGAN was originally proposed for object-level image generation from freehand sketches. Theoretically, it can also be used for
the scene-level freehand sketches. pix2pix~\cite{pix2pix2017} is a popular general image-to-image model which is supposed to be applied in all the image translation tasks. 
We therefore use SketchyGAN~\cite{chen2018sketchygan} and pix2pix~\cite{pix2pix2017} as the baseline methods.

Since we have 14081 pairs of \{scene sketch, scene image\} examples, it is intuitive to directly train the pix2pix and SketchyGAN models to learn the mapping from sketches to images. We therefore conducted the experiments on the entities with lower resolutions, e.g., $128\times 128$. We found that the training of either pix2pix or SketchyGAN was prone to mode collapse, often after 60 epochs (80 epochs for SketchyGAN), even all the 14081 pairs of \{scene sketch, scene image\} examples from the SketchyCOCO dataset were used. The reason may be that the data variety is too huge to be modeled. Even the size of 14K pairs is still insufficient to complete a successful training. However, even with $80\%$ the 14081 pairs of \{foreground image \& background sketch, scene image\} examples, we can still use the same pix2pix model for background generation without any mode collapse. 
\CQ{This may be because the pix2pix model in this case avoids the challenging mapping between the foreground sketches and the corresponding foreground image contents. More importantly, the training can converge fast because the foreground image provides sufficient prior information and constraints for background generation.}  

\noindent\textbf{Comparison with other systems.}
We also compare our approach with the advanced approaches which generate images using constraints from other modalities.
\begin{itemize}
\item \noindent\textbf{GauGAN~\cite{park2019semantic}}: The original GauGAN model takes the semantic maps as input. We found that the GauGAN model can also be used as a method to generate images from semantic sketches where
the edges of the sketches
have category labels as shown in the 7th column of Fig.~\ref{fig:scene_comparison}. In our experiments, we test the public model pre-trained on the dataset COCO Stuff. In addition, we trained a model by taking as input the semantic sketches on our collected SketchyCOCO dataset. The results are shown in Fig.~\ref{fig:scene_comparison} columns 6 and 8.


\item \noindent\textbf{Ashual \emph{et al.}~\cite{ashual2019specifying}}: the approach proposed by Ashual et al. can use either layouts or scene graphs as input. We therefore compared both of the two modes with their pre-trained model. \CQ{To ensure fairness, we test only the categories included in the SketchyCOCO dataset and set the parameter of the minimal object number to 1.} The results are shown in Fig.~\ref{fig:scene_comparison} columns 2 and 4.
\end{itemize}

\noindent\textbf{Qualitative results.} 
From Fig.~\ref{fig:scene_comparison}, we can see the images generated by freehand sketches are much more realistic than those generated from scene graphs or layouts by Ashual \emph{et al.}~\cite{ashual2019specifying}, especially in the foreground object regions. This is because freehand sketches provide a harder constraint compared to scene graphs or layouts (it provides more information including the pose and shape information than scene graphs or layouts). Compared to GauGAN with semantic sketches as input, our approach generally produce more realistic images.
\CQ{Moreover, compared to the GauGAN model trained using semantic maps, our approach also achieves better results, evidence of which can be found in the generated foreground object regions (the cows and elephants generated by GauGAN have blurred or unreasonable textures). }

In general, our approach can produce much better results in terms of the overall visual quality and the realism of the foreground objects than both GauGAN and Ashual \emph{et al.}'s method. The overall visual quality of the whole image is also comparative to the state-of-the-art system. 

\noindent\textbf{Quantitative results.} 
We adopt three metrics to evaluate the faithfulness and realism of the generated scene-level images. 
Apart from FID, the structural similarity metric (SSIM)~\cite{wang2004image} is another metric used to quantify how similar the generated images and the ground truth images are. Higher SSIM value means closer. The last metrics, called FID (local), is used to compute the FID value of the foreground object regions in the generated images. 
From Table~\ref{object_score} we can see most comparison results confirm our observations and conclusions in the qualitative study except for the comparisons with the GauGAN-semantic map model \CQ{and the Ashual et al.~\cite{ashual2019specifying}-layout model in some metrics.}


There are several reasons why the GauGAN model trained using semantic maps is superior to our model in terms of FID and SSIM. Apart from the inherent advantages offered by the semantic map data as a tighter constraint, the GauGAN model trained using the semantic maps contains all the categories in the COCO Stuff dataset, while our model sees only 17 categories in the SketchyCOCO dataset. Therefore, the categories and number of instances in the image generated by GauGAN are the same with ground truth, while our results can contain only a part of them.
\CQ{The Ashual et al.~\cite{ashual2019specifying}-layout model is superior to ours in terms of FID and SSIM. This may be because the input layout information can provide a more explicit spatial constraint than sketches when generating the background. However, our method has greater advantages on the metric of FID (local), which confirms our observation in the qualitative result analysis-that is, our method can generate more realistic foreground images.} 
Because our approach takes as input the freehand sketches, which may be much more accessible than the semantic maps used by GauGAN, we believe that our approach might still be a competitive system for an image-generation tool compared to the GauGAN model.


\CQ{\subsection{Human Evaluation}
We carry out a human evaluation study for both object-level and scene-level results. As shown in Table~\ref{object_score}, we evaluate
the realism and faithfulness of the results from eight object-level and five scene-level comparison models. We select 51 sets of object-level test samples and 37 sets of scene-level test samples, respectively. In the realism evaluation, 30 participants are asked to pick out the resulting image that they think is most ``realistic" from the images generated by the comparison models for each test sample. For the faithfulness evaluation, we conduct the evaluation following SketchyGAN~\cite{chen2018sketchygan} for eight object-level comparison models. Specifically, with each sample image, the same 30 participants see six random sketches of the same category, one of which is the actual input/query sketch. The participants are asked to select the sketch that they think prompts the output image. For five scene-level comparison models, the 30 participants are asked to rate the similarity between the GT image and the resulting images 
on a scale of 1 to 4, with 4 meaning very satisfied and 1 meaning very dissatisfied. In total, $51 \times 8 \times 30 = 12,240$ and $51 \times 30 = 1,530$ trails are respectively collected for object-level faithfulness and realism evaluations, and $37 \times 5 \times 30 = 5,550$ and $37 \times 30 = 1,110$ trails are respectively collected for scene-level faithfulness and realism evaluations.}

\CQ{The object-level statistic results in Table~\ref{object_score} 
generally confirm the quantitative results of faithfulness. The scene-level evaluation shows that our method has the best score on realism, which is not consistent with the quantitative results measured by FID.
This may be because the participants care more about the visual quality of foreground objects than that of background regions. In terms of scene-level faithfulness, GauGAN is superior to our method because the input semantic map generated from the ground truth image provides more accurate
constraints.}

\subsection{Ablation Study}
\label{subsec:abla} 
We conduct comprehensive experiments to analyze each component of our approach, which
includes: a) whether the encoder $E$ has learnt the high level cross-domain attribute information, b) how the joint discriminator $D_J$ works, and c) which GAN model suits our approach the most, and d) whether multi-scale discriminators can be used to improve the results. Due to the limited space, in this section we only present our investigation towards the most important study, i.e., study a) and put the other studies into the supplementary materials. 
\begin{figure}[h]
	\centering
	\includegraphics[width=1.0\linewidth]{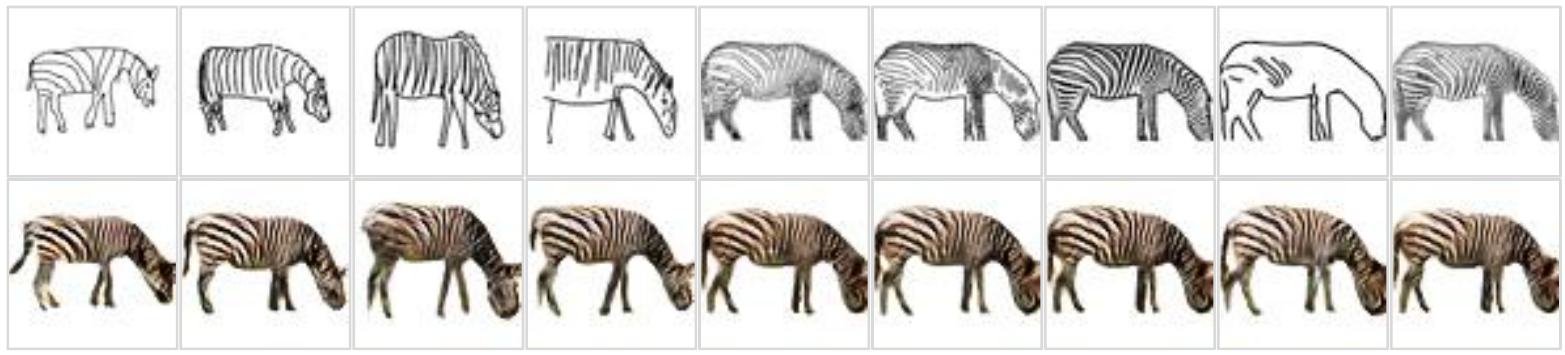}
	\caption{Results from edges or sketches with different style. Column 1 to 4: different freehand sketches. Column 5 to 9: edges from canny, FDoG~\cite{jiang2017configurable}, Photocopy (PC), Photo-sketch~\cite{eitz2009photosketch} and XDoG.~\cite{winnemoller2012xdog}}
	\label{fig:variousSketch_comparison}
\end{figure}

We test different styles of drawings, including sketches and edge maps generated by various filters as input. We show the results in Fig.~\ref{fig:variousSketch_comparison}. We can see that our model works for a large variety of line drawing styles although some of them are not included in the training dataset. We believe that the attribute vector from the Encoder $E$ can extract the high-level attribute information of the line drawings no matter what styles they are.

\section{Discussion and Limitation}
\begin{figure}[h]
	\centering
	\includegraphics[width=1.0\linewidth]{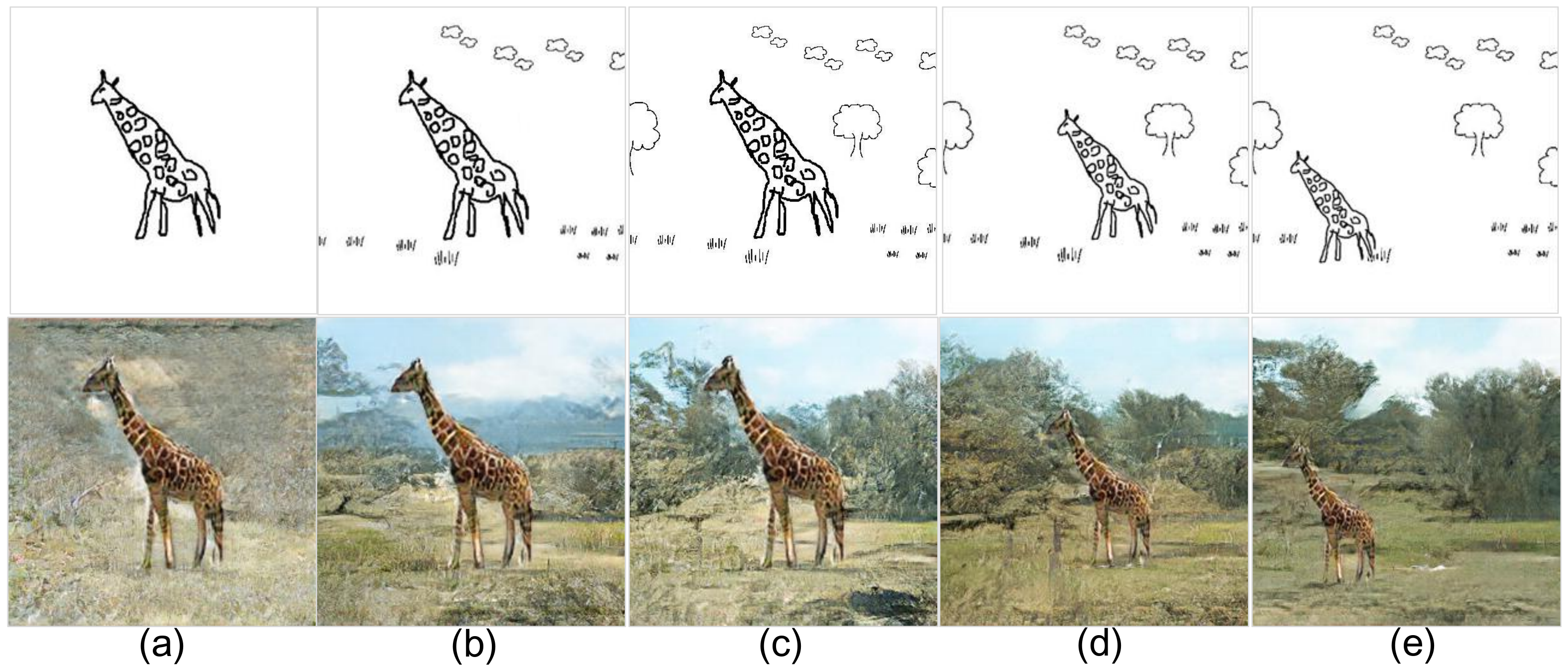}
	\caption{From top to bottom: input sketches, and the images generated by our approach. 
	}
\label{fig:discussion}
\end{figure}

\CQ{\noindent\textbf{Background  generation.}}
We study the controllability and robustness of background generation. As shown in Fig.~\ref{fig:discussion} (a) to (c), we progressively add background categories to the blank background. As a result, the output images are changed reasonably according to the newly added background sketches, which indicates these sketches do control the generation of different regions of the image. It can be seen that although there is 
a large unconstrained blank in the background, the output image is still reasonable. 
We study our approach’s capability of producing diverse results. As shown in Fig.~\ref{fig:discussion}  (c) to (e), we change the location and size of the foreground object in the scene sketch while keeping the background unchanged. As a result, there are significant changes in the background generation. Taking the foreground as a constraint for background training, the foreground and background blend well. We can see the approach even generates shadow under the giraffe. 

\begin{figure}[]
    \centering
	\includegraphics[width=1.0\linewidth]{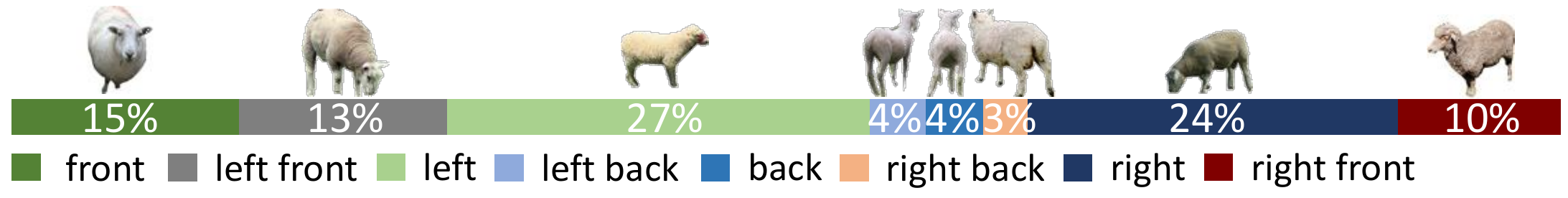}
	\caption{Statistical results of the view angles of foreground objects in SketchyCOCO.}
	\label{fig:data_pose}
\vspace{-5mm}
\end{figure}

\CQ{\noindent\textbf{Dataset Bias.} 
In the current version of SketchyCOCO, all the foreground images for object-level training are collected from the COCO-Stuff dataset. We discard only the foreground objects with major parts occluded from COCO-Stuff in the data collection phrase. To measure the view diversity of the foreground objects, we randomly sample 50 examples from each class in the training data and quantify the views into eight ranges according to the view angles on the x-y plane. This result is shown in Fig.~\ref{fig:data_pose}. As we can see, there are some dominant view angles, such as the side views. We are considering augmenting SketchyCOCO to create a more balanced dataset. 


}

\CQ{\noindent\textbf{Sketch Segmentation.} 
We currently employ the instance segmentation algorithm in \cite{ZouMGDF19} in the instance segmentation step of the scene sketch. Our experiment finds that the adopted segmentation algorithm may fail to segment some objects in the scene sketches in which the object-level sketches are too abstract. To address this problem, we are considering tailoring a more effective algorithm for the task of scene sketch segmentation in the future.}
\section{Conclusion}
For the first time, this paper has presented a neural network based framework to tackle the problem of generating scene-level images from freehand sketches. We have built a large scale composite dataset called SketchyCOCO based on MS COCO Stuff for the evaluation of our solution. Comprehensive experiments demonstrate the proposed approach can generate realistic and faithful images from a wide range of freehand sketches. 

\section*{Acknowledgement}
We thank all the reviewers for their valuable comments and feedback. We owe our gratitude to Jiajun Wu for his valuable suggestions and fruitful discussions that leads to the EdgeGAN model.
This work was supported by the Natural Science Foundation of Guangdong Province, China (Grant No. 2019A1515011075), National Natural Science Foundation of China (Grant No. 61972433, 61921006). 

{\small
\bibliographystyle{ieee_fullname}
\bibliography{egbib}
}

\clearpage
\section*{Supplementary Material}
\setcounter{section}{0}
\section{Objective Function}
Let $\tilde{s}$ be the output edge-image pair and $s$ be the real edge-image pair, $z$ be the noise vector, and $\hat{s}$ be the random sample. Based on our preliminary results, we leverage WGAN-GP as the basis in our network model to achieve stable and effective training. The loss function of WGAN-GP is defined as follows:
\begin{footnotesize}
    \begin{equation}
    \mathcal{L}_{D_{J}}(D)=\underset{\tilde{s} \sim \mathbb{P}_g}{\mathbb{E}} [D_{J}(\tilde{s})] - \underset{s \sim \mathbb{P}_r}{\mathbb{E}} [D_{J}(s)] + \lambda \underset{\hat{s} \sim \mathbb{P}_{\hat{s}}}{\mathbb{E}}[(||\nabla_{\hat{s}} D_{J}(\hat{s})||_2-1)^2].
    \label{eq:GAN_loss_J}
    \end{equation}
\end{footnotesize}
\vspace{-1em}
\begin{footnotesize}
    \begin{equation}
    \mathcal{L}_{D_{J}}(G)=\underset{\tilde{s} \sim \mathbb{P}_g}{\mathbb{E}} [-D_{J}(\tilde{s})].
    \label{eq:GAN_loss_J_G}
    \end{equation}
\end{footnotesize}

Let $\tilde{x}$, $x$ and $\hat{x}$ be the generated edge, real edge and random generated edge, and $\tilde{y}$, $y$ and $\hat{y}$ be the generated natural image, real image and random generated natural image, respectively. Since the discriminators $D_E$ and $D_I$ adopt the same architecture as $D_J$, we can define their losses as:
\begin{footnotesize}
    \begin{equation}
    \mathcal{L}_{D_{E}}(D)=\underset{\tilde{x} \sim \mathbb{P}_g}{\mathbb{E}} [D_{E}(\tilde{x})] - \underset{x \sim \mathbb{P}_r}{\mathbb{E}} [D_{E}(x)] + \lambda \underset{\hat{x} \sim \mathbb{P}_{\hat{x}}}{\mathbb{E}}[(||\nabla_{\hat{x}} D_{E}(\hat{x})||_2-1)^2].
    \label{eq:GAN_loss_E}
    \end{equation}
\end{footnotesize}
\vspace{-1em}
\begin{footnotesize}
    \begin{equation}
    \mathcal{L}_{D_{E}}(G)=\underset{\tilde{x} \sim \mathbb{P}_g}{\mathbb{E}} [-D_{E}(\tilde{x})].
    \label{eq:GAN_loss_E_G}
    \end{equation}
\end{footnotesize}
\begin{footnotesize}
    \begin{equation}
    \mathcal{L}_{D_{I}}=\underset{\tilde{y} \sim \mathbb{P}_g}{\mathbb{E}} [D_{I}(\tilde{y})] - \underset{y \sim \mathbb{P}_r}{\mathbb{E}} [D_{I}(y)] + \lambda \underset{\hat{y} \sim \mathbb{P}_{\hat{y}}}{\mathbb{E}}[(||\nabla_{\hat{y}} D_{I}(\hat{y})||_2-1)^2].
    \label{eq:GAN_loss_I}
    \end{equation}
\end{footnotesize}
\vspace{-1em}
\begin{footnotesize}
    \begin{equation}
    \mathcal{L}_{D_{I}}(G)=\underset{\tilde{y} \sim \mathbb{P}_g}{\mathbb{E}} [-D_{I}(\tilde{y})].
    \label{eq:GAN_loss_I_G}
    \end{equation}
\end{footnotesize}

During training, $D_J$, $D_E$ and $D_I$ are updated to minimize Equation \ref{eq:GAN_loss_J}, Equation \ref{eq:GAN_loss_E} and Equation \ref{eq:GAN_loss_I} separately. Our classifier is used to predict class labels. We use the focal loss~\cite{lin2018focal} as the classification loss. Let $c$ be the predicted class label. Formally, the loss function between the ground-truth label and the predicted label is defined as:
\begin{footnotesize}
    \begin{equation}
    \mathcal{L}_{ac}(D)=\mathbb{E}[log P(C=c | y)]. 
    \label{eq:focal_loss_D}
    \end{equation}
\end{footnotesize}
Classifier is trained to maximize Equation \ref{eq:focal_loss_D}. The generator also maximizes $\mathcal{L}_{ac}(G)=\mathcal{L}_{ac}(D)$ when the classifier is fixed. 

We train the encoder $E$ with the $L1$ loss between the random input vector $z$ and generated vector $\tilde{z}$. $\tilde{z}$ is from encoding the edge $\tilde{x}$ which is the output of the generator $G_E$. Formally, the loss function is defined as:
\begin{footnotesize}
    \begin{equation}
    \mathcal{L}_1^{ \text{latent}}(E)=\underset{z \sim \mathbb{P}_z}
    {\mathbb{E}}||z-E(\tilde{x})||_1.
    \label{eq:L1_loss}
    \end{equation}
\end{footnotesize}
In summary, the loss function of the generator $G_E$ is:
\begin{footnotesize}
    \begin{equation}
    \mathcal{L}_{G_{E}}(G)=\mathcal{L}_{D_{J}}(G)+\mathcal{L}_{D_{E}}(G).
    \label{eq:loss_G_E}
    \end{equation}
\end{footnotesize}
and the loss function of the generator $G_I$ is:
\begin{footnotesize}
    \begin{equation}
    \mathcal{L}_{G_{I}}(G)=\mathcal{L}_{D_{J}}(G)+\mathcal{L}_{D_{I}}(G)-\mathcal{L}_{ac}(G).
    \label{eq:loss_G_I}
    \end{equation}
\end{footnotesize}
The generator $G_E$ minimizes Equation \ref{eq:loss_G_E} and $G_I$ minimizes Equation \ref{eq:loss_G_I}.
\section{Implementation Details}
In the stage of instance generation, we train the model with 100 epochs, and randomly generate latent vectors in the normal distribution with zero mean and variance of 1.0. 
We train the generator and the discriminators with instance normalization. The encoder is implemented with ResNet blocks using instance normalization and ReLU, and the classifier is implemented with MRU block~\cite{chen2018sketchygan}. We use ReLU and Tanh for the generator while Leaky ReLU for the discriminator. In the instance of DCGAN, we use the Adam optimizer with a learning rate of 0.0002 and a beta of 0.5 for all the networks. In the instance of WGAN~\cite{arjovsky2017wasserstein}, we clamp the weights between $-$0.01 and 0.01 after each gradient update and use the RMSprop optimizer with a learning rate of 0.0002 for all the networks. In the instance of WGAN-GP~\cite{gulrajani2017improved}, we set the weight of gradient penalty $\lambda$ to 10, using the RMSprop optimizer with a learning rate of 0.0002 for all the networks. For background generation, we train the pix2pix model with the 110 epochs. We use XDoG~\cite{winnemoller2012xdog} to obtain the edge maps of objects for training data.
\section{Representative Samples from SketchyCOCO}
We show more examples of SketchyCOCO including five-tuple ground truth data in Fig.~\ref{fig:SketchyCOCO}.

\begin{figure*}[h]
	\centering
	\includegraphics[width=1.0\linewidth]{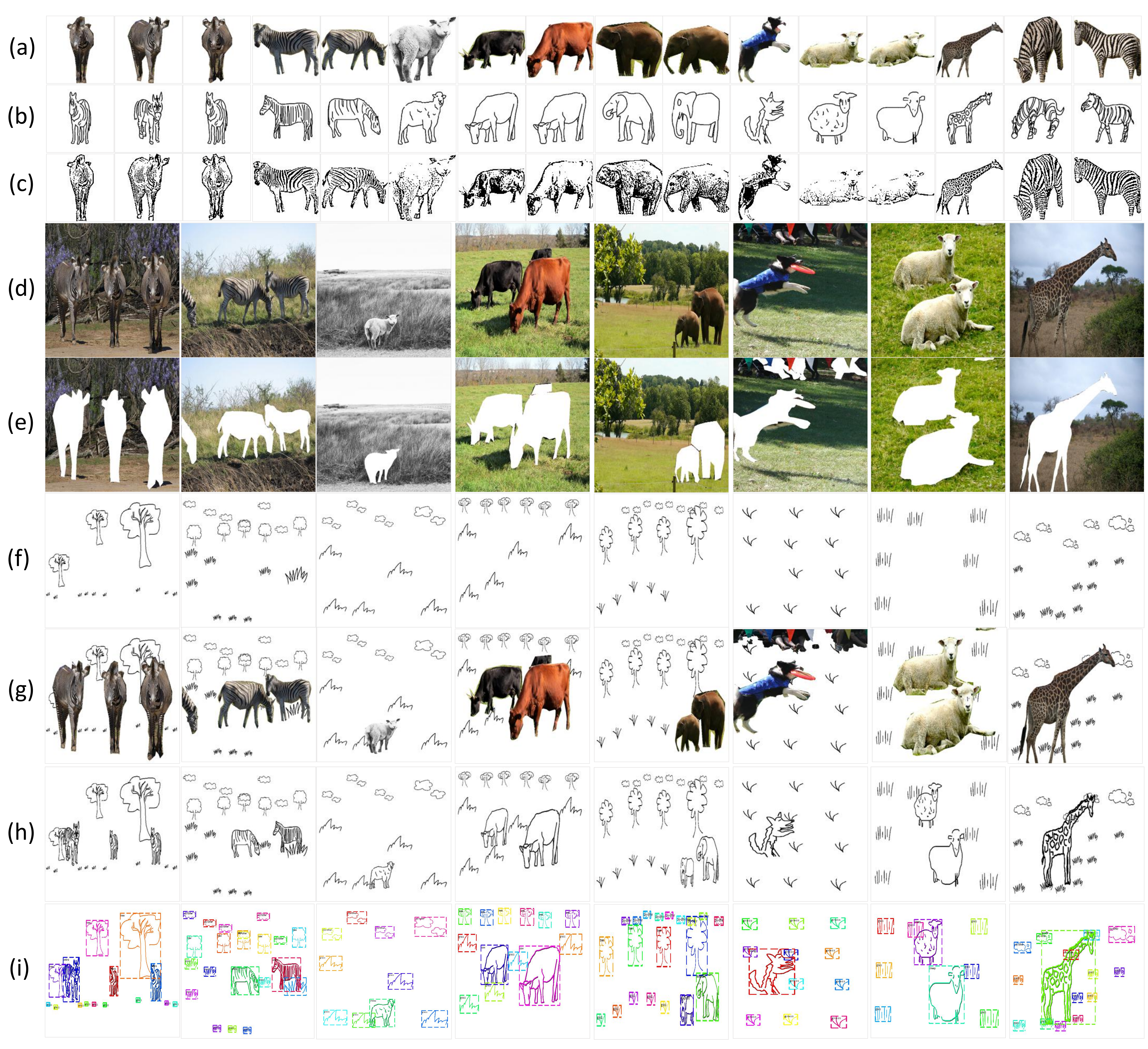}
	\caption{More exmples of the five-tuple ground truth data of SketchyCOCO, i.e., (1) \{foreground image(a), foreground sketch(b), foreground edge maps(c)\}, (2) \{background image(e), background sketch(f)\}, (3) \{scene image(d), foreground image \& background sketch(g)\}, (4) \{scene image(d), scene sketch(h)\}, and (5) sketch segmentation(i). Unlike foreground sketches depicting single objects, background sketches such as grass and trees are purposefully designed to depict a specific region (e.g., several tree sketches depict a forest).}
	\label{fig:SketchyCOCO}
\end{figure*}
\section{Object Level Results}
\begin{figure}[h]
	\centering
	\includegraphics[width=1.0\linewidth]{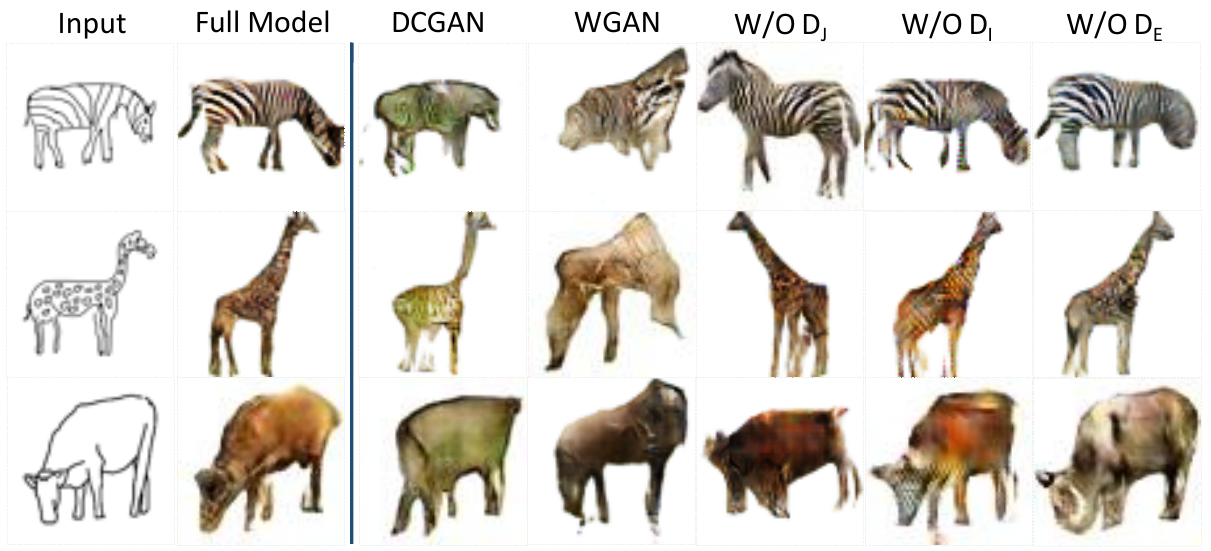}
	\caption{The results of network ablation. The full model is based on WGAN-gp and contains $D_J$, $D_I$ and $D_E$. WGAN and DCGAN are the structures replacing WGAN-gp with WGAN and DCGAN, respectively.}	
	\label{fig:network_ablation}
\end{figure}

\begin{figure*}[h]
	\centering
	\includegraphics[width=1.0\linewidth]{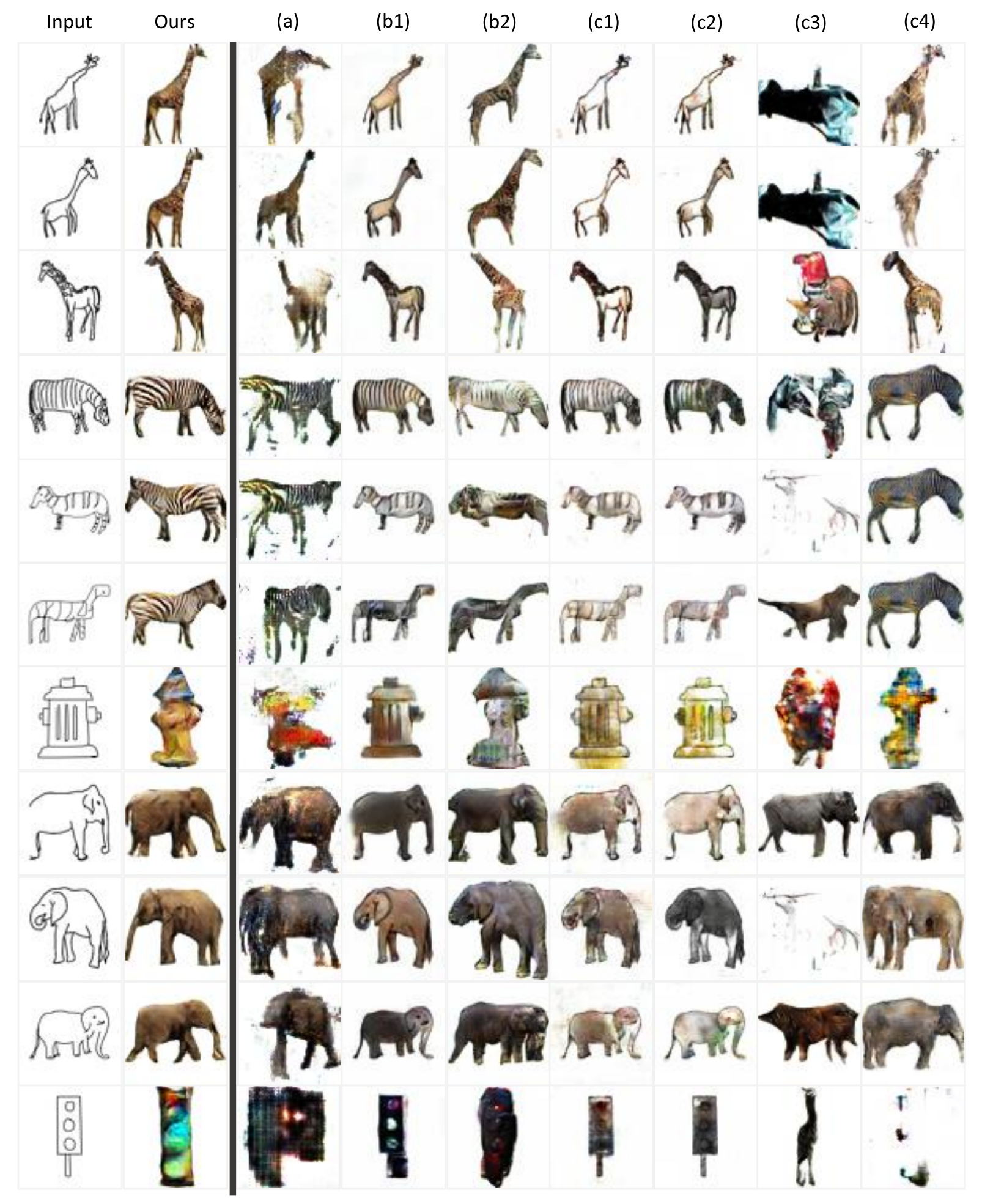}
\end{figure*}

\begin{figure*}[h]
	\centering
	\includegraphics[width=1.0\linewidth]{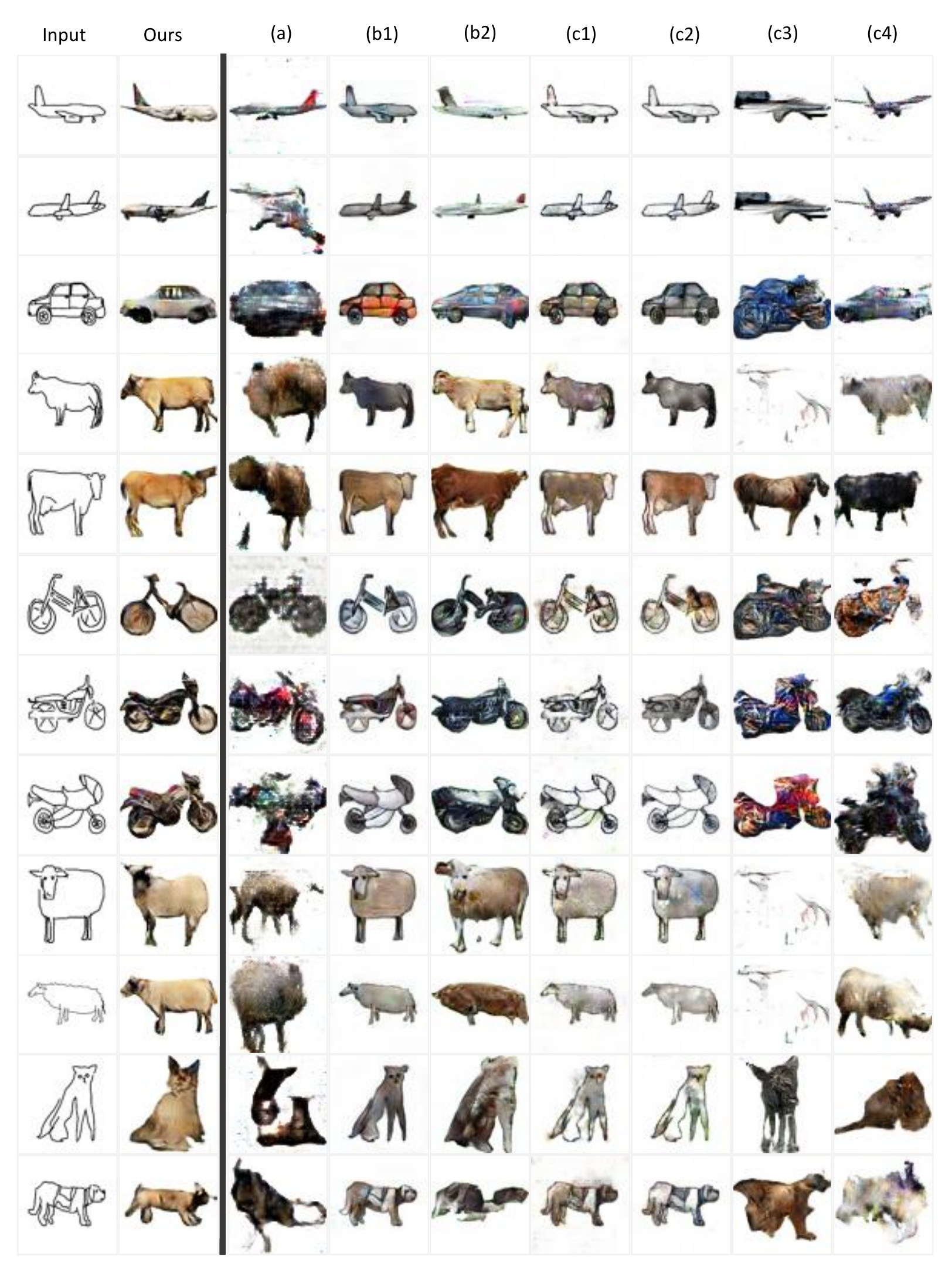}
\end{figure*}

\begin{figure*}[h]
	\centering
	\includegraphics[width=1.0\linewidth]{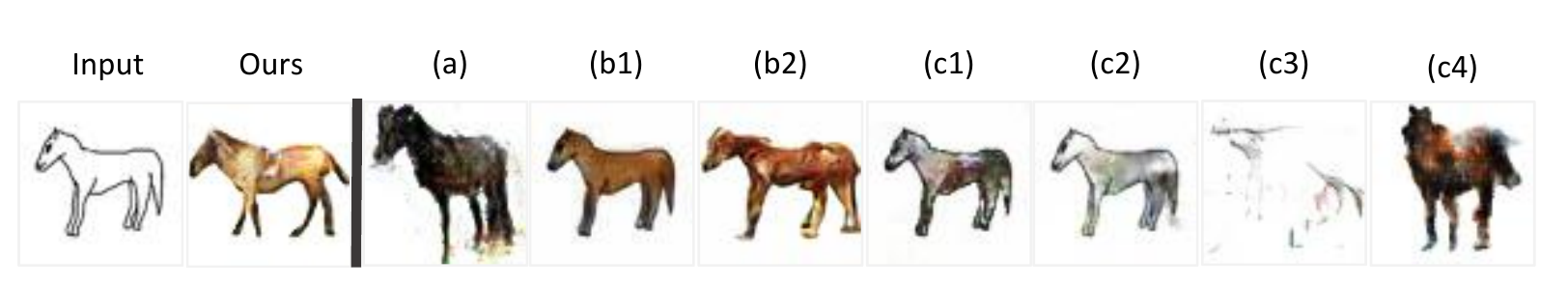}
	\caption{From left to right: input sketches, results from edgeGAN, ContextualGAN~(a), two training modes of SketchyGAN (i.e., SketchyGAN-E~(b1) and SketchyGAN-E\&S)~(b2), and four training modes of pix2pix (i.e.,  pix2pix-E-SEP~(c1),  pix2pix-E-MIX~(c2), pix2pix-S-MIX (c3), and pix2pix-S-SEP (c4)).}	
	\label{fig:object_comparison}
\end{figure*}

\begin{figure*}[h]
	\centering
	\includegraphics[width=0.95\linewidth]{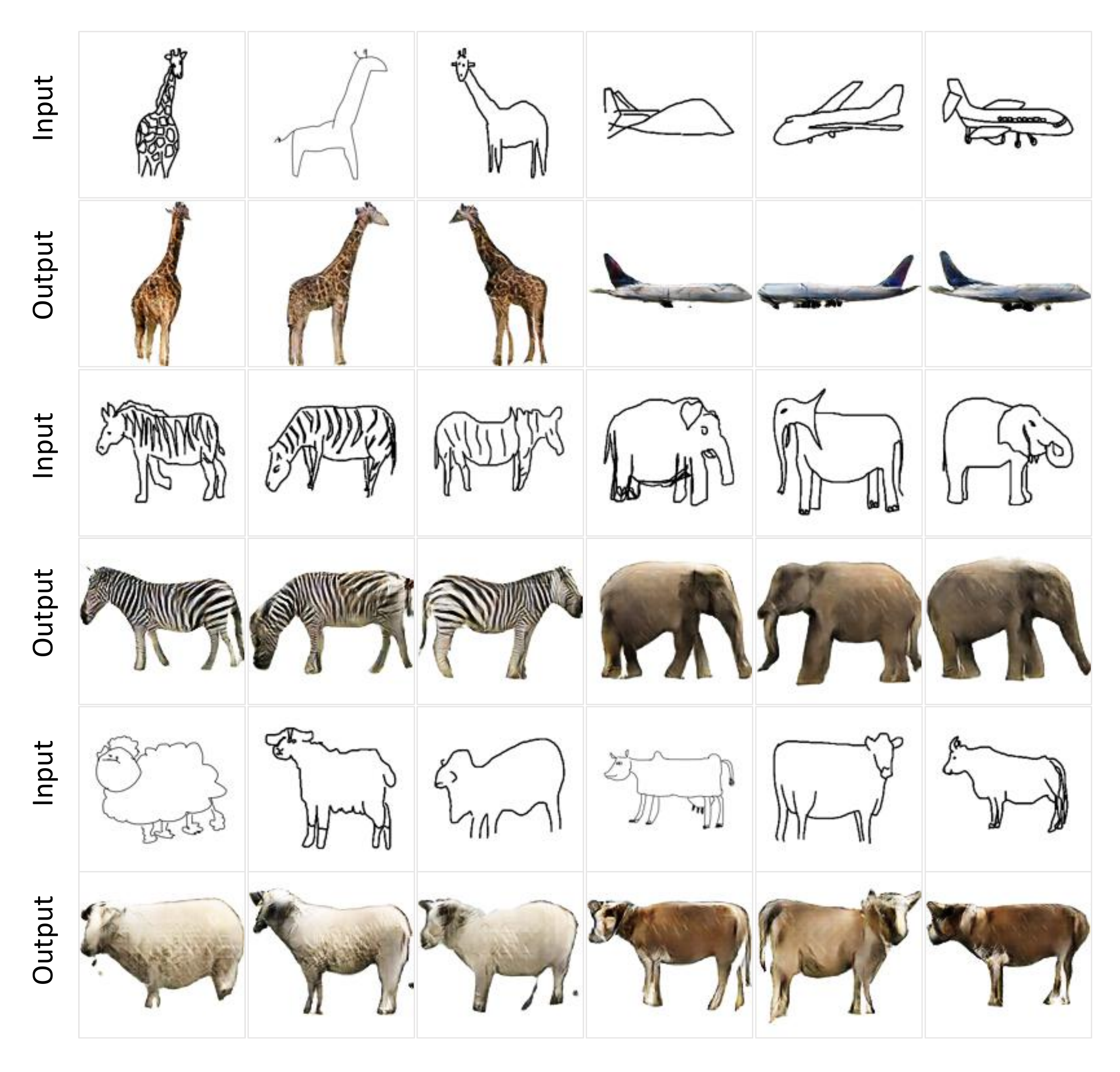}
	\caption{More $128\times 128$ results in the object level.}	
	\label{fig:object_128}
\end{figure*}

\subsection{More object level comparison results}
We compare edgeGAN with ContextualGAN~\cite{lu2018image}, SketchyGAN~\cite{chen2018sketchygan}, and pix2pix~\cite{pix2pix2017} under different training strategies. Fig.~\ref{fig:object_comparison} shows the comparison results. This figure is a supplement to Fig. 6 in the paper.

\subsection{Some $128 \times 128$ results in the object level}
We have trained the model on images of resolution $128 \times 128$. What is different from training on images of resolution $64 \times 64$ is that we use one more discriminator, whose structure is a copy of $D_I$. And we set the size of its input to  $64 \times 64$ to guarantee the global information. In addition, we also set the size of $D_I$'s input to $256 \times 256$ so that the model can pay more attention to local details. Some results are shown in Fig.~\ref{fig:object_128}.
\section{Scene Level Results}

\begin{figure*}[h]
	\centering
	\includegraphics[width=1.0\linewidth]{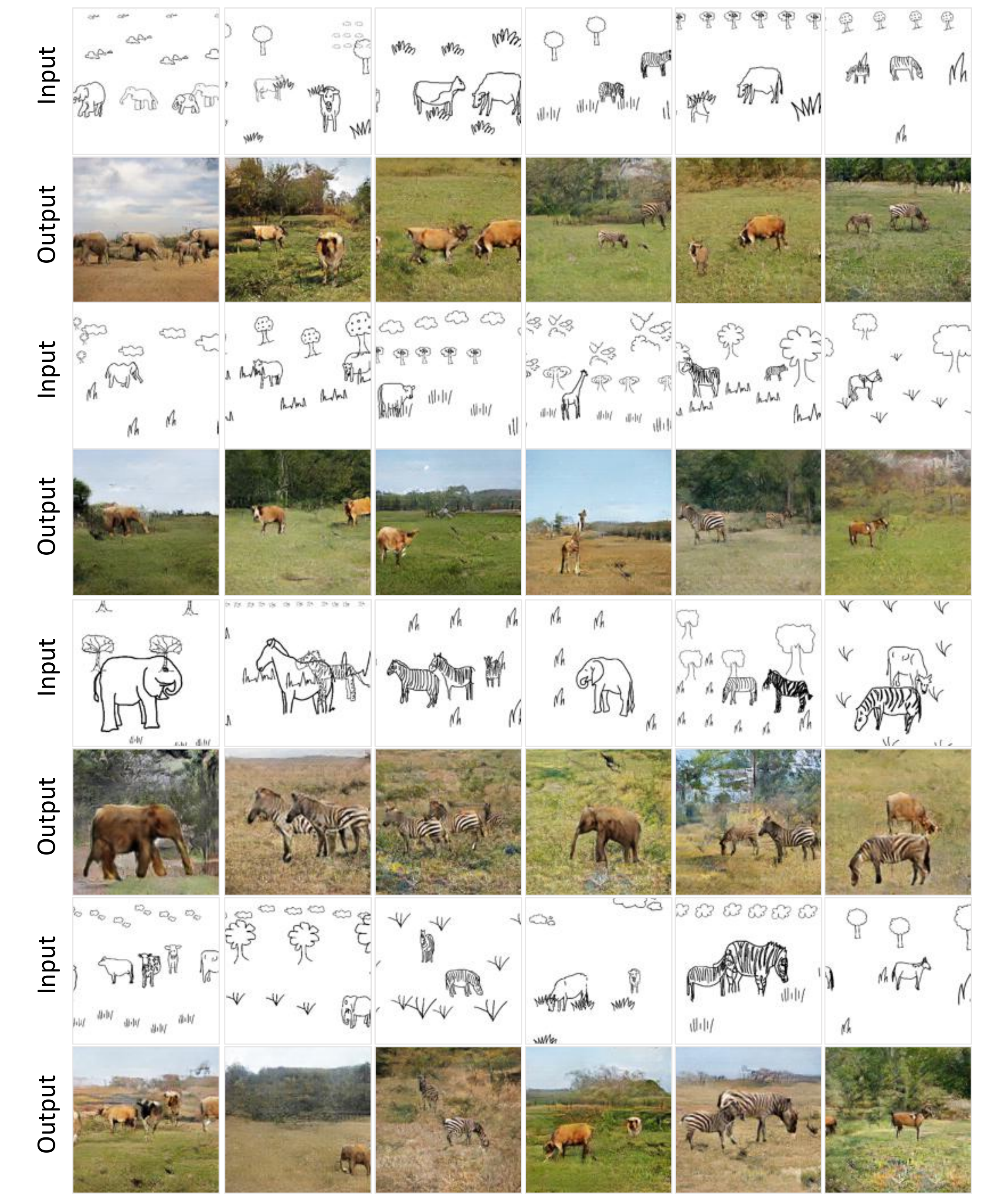}
	\caption{More $128 \times 128$ results in the scene level.}
	\label{fig:scene_128}
\end{figure*}

\begin{figure*}[h]
	\centering
	\includegraphics[width=1.0\linewidth]{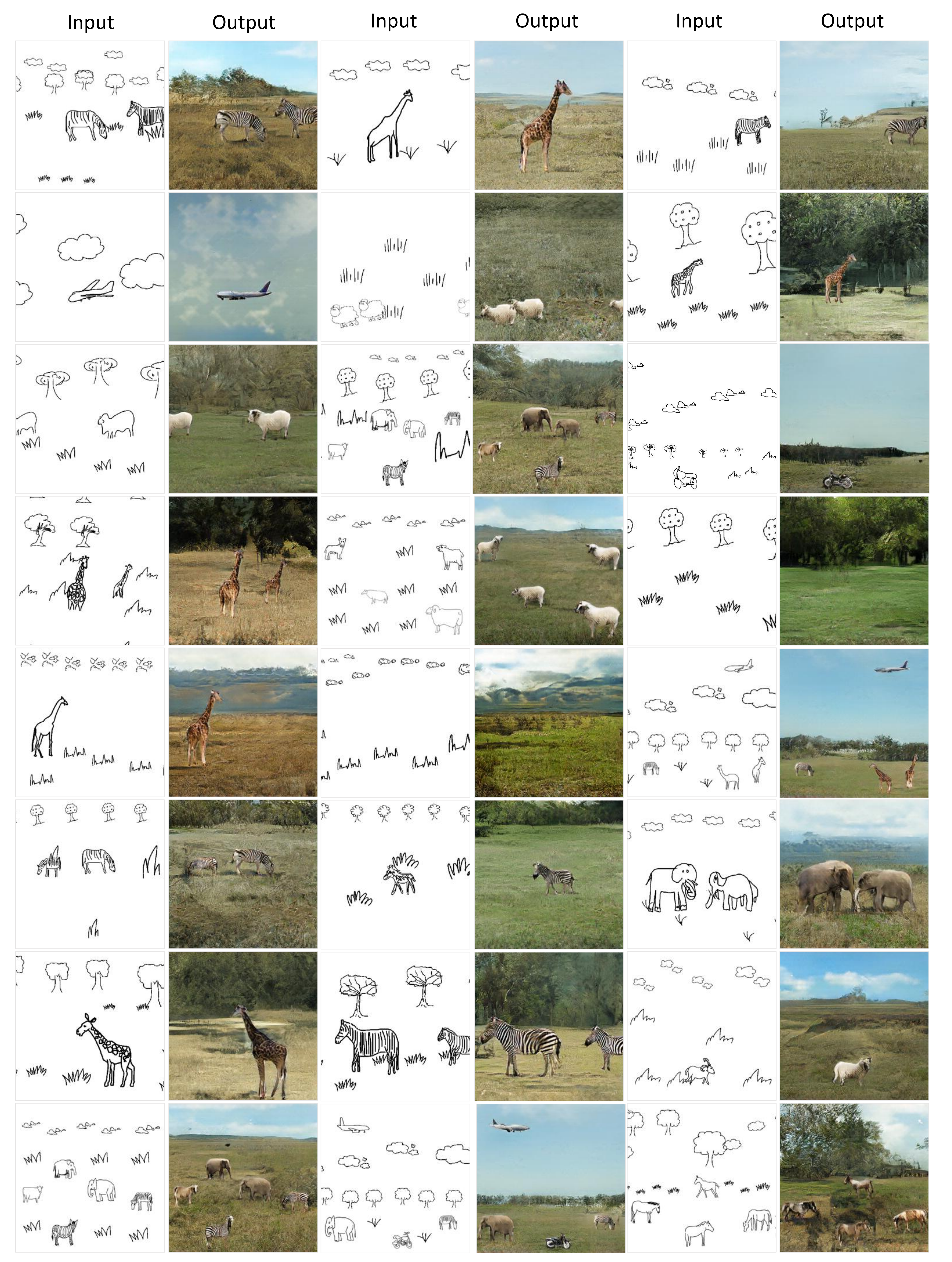}
\end{figure*}

\begin{figure*}[h]
	\centering
	\includegraphics[width=1.0\linewidth]{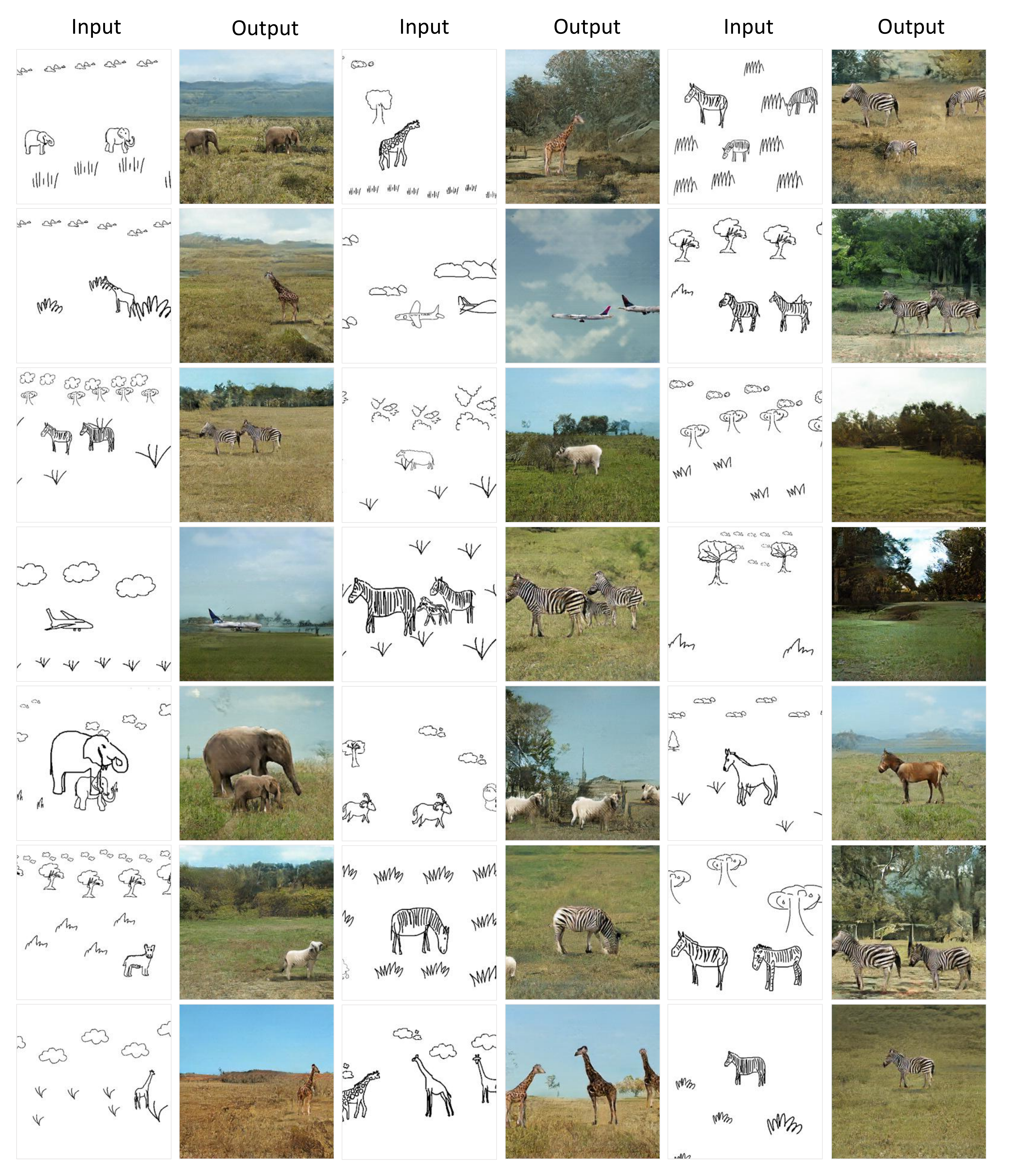}
	\caption{More $256 \times 256$ results in the scene level.}
	\label{fig:scene_256}
\end{figure*}

In this section, we show more $128 \times 128$ scene-level results
in Fig.~\ref{fig:scene_128}, which are generated based on the $64 \times 64$ object level results, as well as more $256 \times 256$ results in Fig.~\ref{fig:scene_256}, which are generated based on the $128 \times 128$ object level results.
\section{Ablation Study}

\begin{table}
\caption{Object level scores in the ablation study.}
\begin{center}
	\begin{tabular}{c c c c}
		\hline
		Model(object) & FID & Accuracy & Shape Similarity\\
		\hline
		Full Model & \textbf{87.59} & \textbf{0.8866} & \textbf{2.294e+04}\\
		W/O $D_{J}$ & 95.63 & 0.8361 &  2.457e+04\\
		W/O $D_{I}$ & 110.17 & 0.6964 & 2.331e+04\\
		W/O $D_{E}$ & 91.12 & 0.8204 & 2.341e+04\\
		DCGAN & 108.86 & 0.6429 & 2.335e+04\\
		WGAN & 106.67 & 0.3172 & 2.471e+04\\
		\hline
	\end{tabular}
\label{object_score}
\end{center}
\end{table}

\begin{itemize}
\item \noindent\textbf{How the joint discriminator $D_J$ works?} 
We concat the outputs of the edge generator and the image generator in the width channel as a joint image, which is used as the fake input of $D_J$. The real edge-image pair is taken as the real input. Therefore, the generated edge and image from the same vector respect each other under the constraint of the adversarial loss. In the inference stage, the attribute vector, which can be mapped to an edge image close to the input sketch, also can be mapped to a natural image with reasonable pose and shape. As shown in Fig.~\ref{fig:network_ablation}, the pose and shape of the generated image are not correct without $D_J$.  

\item \noindent\textbf{Which GAN model suits our approach the best?}  
WGAN-gp was proved to be more suitable for small data sets than DCGAN and WGAN, making training more stable and producing higher quality results. As shown in Fig.~\ref{fig:network_ablation}, when we change it to DCGAN or WGAN, the results get worse in both faithfulness and realism. So our network is based on WGAN-gp. More quantitative results are shown in Table~\ref{object_score}.

\item \noindent\textbf{Whether multi-scale discriminators can be used to improve the results?}
We use multi-scale discriminators to improve the quality of generated images. For resolution $64 \times 64$, we add an edge discriminator ($D_E$) and an image discriminator ($D_I$), the inputs of which are  enlarged edge ($128 \times 128$) and image ($128 \times 128$) respectively. As a result, the model can learn  smaller receptive fields and thus pay more attention to local details. As shown in Fig.~\ref{fig:network_ablation}, the quality of local details is not as good as that of the full model when without $D_I$ or $D_E$. As shown in Table~\ref{object_score}, the full model outperforms the model without the multi-scale discriminators on both realism and faithfulness metrics.
\end{itemize}

\end{document}